\documentclass[12pt]{alt2021} 

\title[Online Estimation of Sample Statistics with Comparison Feedback]{Learning with Comparison Feedback:\\Online Estimation of Sample Statistics}
\usepackage{times}
\usepackage{float}

\usepackage{macros}

\author{\Name{Michela Meister} \Email{meister@cs.cornell.edu}\\
\Name{Sloan Nietert} \Email{nietert@cs.cornell.edu}\\
\addr Cornell University}

\begin{document}

\maketitle

\begin{abstract}%
We study an online version of the noisy binary search problem where feedback is generated by a non-stochastic adversary rather than perturbed by random noise. 
We reframe this as maintaining an accurate estimate for the median of an adversarial sequence of integers, $x_1, x_2, \dots$, in a model where each number $x_t$ can only be accessed through a single threshold query of the form ${1(x_t \leq q_t)}$.
In this online comparison feedback model, we explore estimation of general sample statistics, providing robust algorithms for median, CDF, and mean estimation with nearly matching lower bounds. We conclude with several high-dimensional generalizations.
\end{abstract}


\section{Introduction}\label{section:intro}

Imagine that we seek to understand an unknown real distribution $\D$ but are unable to sample from $\D$ directly. Instead, we are allowed comparison queries of the form: is a sample $x \sim \D$ less than $q$? This partial feedback models situations when exact values contain sensitive information or are expensive to obtain. For example, survey participants might more willingly or accurately answer yes-or-no questions about personal details like health habits or income compared to similar questions asking for quantitative values. Likewise, customers frequently answer queries of the form ``is this item worth at least \$q to me'' (by making purchases) while being unaware of their exact valuation of the item.

In this setting, perhaps the most natural statistic of $\D$ to estimate is its median. Letting $X$ denote a random variable with distribution $\D$, the feedback we receive for query $q$ can be expressed as $A_q = 1(X \leq q)$. Note that each $A_q$ is a Bernoulli random variable with expectation $\Exp[A_q]$ equal to $\Pr(X \leq q) = F_X(q)$, i.e., the cdf of $X$ evaluated at $q$. Thus, we have a family of random variables $A_q$ with means increasing in $q$ from 0 to 1, and finding the median of $\D$ amounts to searching for $q$ such that $\Exp[A_q] \approx 1/2$. The discrete case of $n$ Bernoulli random variables was examined as ``noisy binary search'' by \cite{karp2007noisy}, who provided two algorithms to tackle this problem which both require $O(\log n / \varepsilon^2)$ queries for a natural measure of error $\varepsilon$.

\subsection{Problem Description}
We now relax this distributional assumption for the samples, considering the following online extension of the above problem. Suppose that our samples, now taken to be integers between 1 and $n+1$, are generated by an adversary rather than taken i.i.d.\ from a distribution. At each time $t$, the adversary produces a sample $x_t \in [n+1]$, and the algorithm $\A$ generates a query $q_t \in [n]$ --- each ignorant of the other's choice\footnote{We omit $n+1$ from the query set because $1(x_t \leq n+1) \equiv 1$.}. Then, $\A$ receives feedback $1(x_t \leq q_t)$ and produces a median estimate $\hat{m}_t$ of $x_1, \dots, x_t$, while $q_t$ is revealed to the adversary. We consider both oblivious and adaptive adversaries, where an oblivious adversary commits to the sequence $x_1, x_2, \dots$ at the start, while an adaptive adversary may select $x_t$ based on the history of the prior $t-1$ time steps. Let $F_t$ defined by $F_t(i) = \frac{1}{t}\sum_{\tau = 1}^t 1(x_\tau \leq i)$ be the empirical CDF of the sequence $x_1, \dots, x_t$, where $F_t(0) = 0$. We define the median estimation error of the algorithm at time $t$ as in \citep{karp2007noisy} by
\begin{align}\label{eq:est-error}
    E(t) = E(\hat{m}_t) &= \min \{ \eps \geq 0 \mid [F_t(\hat{m}_t-1), F_t(\hat{m}_t)] \cap [1/2 - \eps, 1/2 + \eps] \neq \emptyset \}\\
    &= \dist([F_t(\hat{m}_t-1), F_t(\hat{m}_t)], 1/2).\nonumber
\end{align}
An estimate $\hat{m}$ is called $\eps$-good if $E(\hat{m}) \leq \eps$.

We examine the capacity of an algorithm to guarantee low error $E(T)$ at some evaluation time $T$. Both the algorithm and adversary can take either fixed-horizon or anytime forms; a fixed-horizon agent is provided the evaluation time $T$ in advance, whereas an anytime agent must be prepared to continue indefinitely without knowledge of $T$. We always assume a stronger, fixed-horizon adversary, since this is no obstacle to our upper bounds and is proven to be without loss of generality for lower bounds in \cref{section:preliminaries}. To quantify an algorithm's performance against a class of adversaries $\mathfrak{B}$, we more formally write $E(T,\A,\mathcal{B})$ to denote the random error of algorithm $\A$ at time $T$ against adversary $\mathcal{B} \in \mathfrak{B}$, where $T$ is given as an initial parameter to $\mathcal{B}$ (and $\A$ in the fixed-horizon case). We then define the query complexity of $\A$ against $\mathfrak{B}$ to be the minimum $T_0 = T_0(n,\eps)$ such that, for all $T > T_0$ and $\mathcal{B} \in \mathfrak{B}$, $E(T,\A,\mathcal{B}) \leq \eps$ with probability at least $3/4$\footnote{We demonstrate in Lemma \ref{appendix:confidence-boosting} that, for the algorithms we present, this confidence probability can be boosted from $3/4$ to $1 - \delta$ at the cost of a multiplicative factor of $\log(1/\delta)$, using a median of medians approach.}.

\begin{figure}[htbp]
\centering
\includegraphics[scale = 0.75]{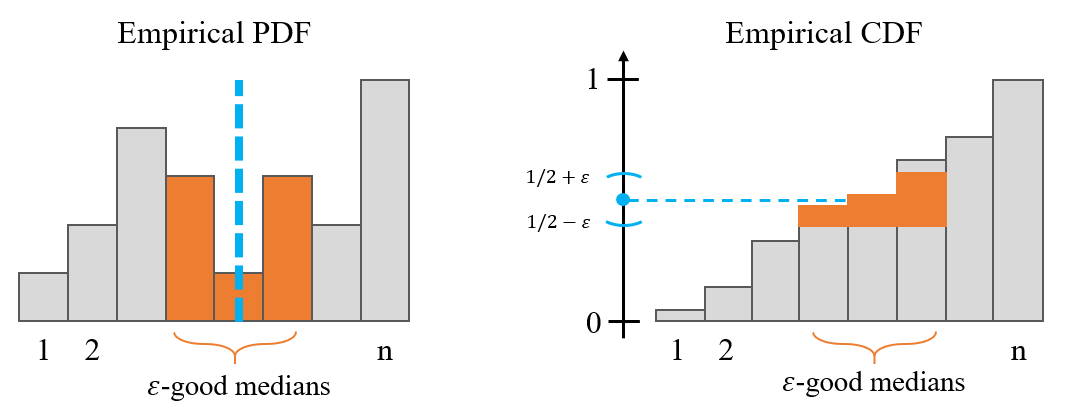}
\caption{Visualization of median estimates $\hat{m}$ with $E(\hat{m}) \leq \eps$}
\end{figure}

We also consider online estimation of the following sample statistics:
\begin{itemize}
    \item arbitrary quantile estimation, where $\A$ maintains an estimate $\hat{m}_t$ of the empirical $\tau$-quantile, for $\tau \in [0,1]$, and $1/2$ is substituted with $\tau$ in \eqref{eq:est-error};
    \item full CDF estimation, where $\A$ maintains an estimate $\hat{F}_t:[n] \to [0,1]$ of the empirical CDF and error $E(t)$ is defined as the Kolmogorov–Smirnov distance
    \begin{equation*}
        \|\hat{F}_t - F_t\|_\infty := \sup_{i \in [n]}|\hat{F}_t(i) - F_t(i)|;
    \end{equation*}
    \item mean estimation, where $\A$ maintains a mean estimate $\hat{\mu}_t$ of the empirical mean $\mu_t$ and error is defined as $E(t) = |\hat{\mu}_t - \mu_t|/n$. 
\end{itemize}

\subsection{Summary of Contributions}
After establishing some preliminaries, \cref{section:online} provides a randomized algorithm for online CDF estimation with anytime query complexity $O(n \log (n)/\eps^2)$ against even adaptive adversaries. This procedure is easily modified to obtain a median estimation algorithm with the same guarantees. Next, we show that both estimation algorithms have essentially optimal query complexities in general, with information theoretic lower bounds. We further observe that randomness is essential in the adversarial setting, proving that any deterministic median estimation algorithm incurs constant error. In \cref{section:stochastic}, we explore the regime of $\eps = O(1)$, where potential improvement in still possible, providing an algorithm for stochastic CDF estimation with query complexity logarithmic in $n$ that is not addressed by our current lower bounds.
\cref{section:means} examines the simpler problem of mean estimation, giving an algorithm with anytime query complexity $1/\eps^2$ against even an adaptive adversary. Further, we show that no estimation algorithm can obtain lower mean squared error.
Finally, we propose some intriguing generalizations of this online comparison feedback model, considering graphs, higher-dimensional spaces, and online convex programming.

\subsection{Related Work}
A generalization of noisy binary search is the ``noisy twenty questions'' game proposed by ~\cite{renyi1961}, where the goal is to identify an element in a set from noisy or otherwise faulty responses to comparison queries. Numerous variations of this problem, and the noisy binary search problem, are surveyed by ~\cite{pelc2002}.

Our work is inspired by the particular noisy binary search problem explored by \cite{karp2007noisy}, where there is a sequence of $n$ Bernoulli random variables with unknown, non-decreasing means $p_1, \dots, p_n$. At each time step, an algorithm samples from one of the random variables, and the objective is to find the index of the random variable with mean closest to $1/2$. The naive solution is essentially binary search, where the algorithm draws sufficient samples from $p_{n/2}$ to determine whether $p_{n/2} < 1/2$, and then recurses on the appropriate interval. This simple procedure requires $O(\log(n)\log\log(n)/\eps^2)$ samples in expectation, and Karp and Kleinberg give two more efficient algorithms, one via multiplicative weights and one via backtracking binary search, which each require only $O(\log(n)/\eps^2)$ samples.

Later work by \cite{ben2008bayesian} and \cite{braverman2008} explores the related problems of searching and sorting when comparisons are faulty with some fixed probability. Additionally, \cite{nowak2009noisy} extends the search problem to the so-called generalized binary search setting, where feedback is of the form $h(q_t)$ plus noise for some $h$ in a hypothesis class $\H$, and the goal is to determine $h$ with as few queries as possible.

In general, one can draw parallels between our setting and various pure exploration problems explored within the multi-armed bandit and experimental design communities. Particularly relevant is work on so-called monotone thresholding bandits explored by \cite{garivier2017}, where bandit arms have increasing means and one must identify that with mean closest to a desired threshold. Our setup can be viewed within this framework, as well as that of \cite{karp2007noisy}, although we reject the stochastic assumptions for rewards/samples. Instead we suppose that an adversary selects monotone $\{0,1\}$-valued reward/sample realizations at each time step, and the algorithm is still restricted to viewing only the realization for its selected arm/query.

Computation with noisy comparison feedback has also been studied on decision trees and in the context of searching on a binary tree by ~\cite{feige1994}. ~\cite{kempe2016} examine the query complexity of noisy search on graphs, where the goal is to identify a target vertex $v$. In this model, an agent is allowed to make a vertex query $q$ at each time step, and if $q \neq v$ the feedback received is a neighbor of $q$ along the shortest path to $v$, which may be perturbed by some noise. Working with this same model, ~\cite{dereniowski2020} improve the computational complexity of selecting a vertex to query by approximating the graph median, an object that we briefly discuss in \cref{section:generalizations}.



\section{Preliminaries}\label{section:preliminaries}


Before presenting the main results, we introduce two useful lemmas which simplify our exploration of the online comparison feedback landscape, with proofs reserved for \cref{appendix:preliminaries}. First, we establish that arbitrary quantile estimation is no harder than median estimation, in much the same way as \cite{karp2007noisy}. 

\begin{lemma}[Reduction to $\tau = 1/2$]\label{lemma:median-to-quantile}
Let $\A$ be an online median estimation algorithm with query complexity $T_0(n,\eps)$ against some class of adversaries in the comparison feedback model. Then, for any $\tau \in [0,1]$, there exists a modified algorithm $\A'$ which maintains a $\tau$-quantile estimate with query complexity $\max\{T_0(n,\eps/2), \Theta(1/\eps^2)\}$ in the same setting.
\end{lemma}

Next, we justify our restriction to fixed-horizon adversaries, demonstrating that they can be modified to provide similar lower bounds without advanced knowledge of the time horizon.

\begin{lemma}[Reduction to Fixed-Horizon Adversaries]\label{lemma:fixed-horizon-reduction}
Let $\mathcal{B}$ be a fixed-horizon online adversary which forces an online median, CDF, or mean estimation algorithm $\A$ to admit error $E(T) > \eps$ with probability greater than $1/4$ for an infinite sequence of time horizons $T$ beginning at $T_0$. Then, there exists a modified anytime adversary $\mathcal{B}'$ which does the same (where the sequence of horizons, while still beginning at $T_0$, may be distinct from that of $\mathcal{B}$).
\end{lemma}

\section{Online CDF and Median Estimation}\label{section:online}

We first examine online estimation of CDFs and medians within our model, providing both algorithms and lower bounds.

\subsection{Upper Bounds}\label{subsection:upper_bounds}
The presented algorithms succeed against even an adaptive adversary and have error bounds guaranteed to hold at every round, not just at the time horizon. Their behavior is quite simple -- both query uniformly at random -- though we pay for this with a linear dependence on $n$.

As stated, our algorithm $\textsc{CdfEst}$ generates i.i.d.\ queries $q_1, q_2, \dots$ sampled uniformly from $[n]$. At each time $t$, with cumulative feedback $1(x_1 \leq q_1), \dots, 1(x_t \leq q_t)$, $\textsc{CdfEst}$ computes the CDF estimate $\hat{F}_t$ defined by
\begin{equation}
    \hat{F}_t(i) = \frac{n}{t} \sum_{\substack{\tau \in [t]\\q_\tau = i}} 1(x_\tau \leq q_\tau) = \frac{n}{t} \sum_{\substack{\tau \in [t]\\q_\tau = i}} 1(x_\tau \leq i), \quad \forall i \in [n],
\end{equation}
with $\hat{F}_t(n+1) = 1$. Recall that we defined query complexity for CDF estimation in terms of the Kolmogorov-Smirnov distance $\|\hat{F}_t - F_t\|_\infty = \sup_{i \in [n+1]}|\hat{F}_t(i) - F_t(i)|$.

\begin{theorem}\label{thm:cdf-est-upper-bound}
In the online comparison feedback model, $\textsc{CdfEst}$ achieves anytime query complexity $O(n\log(n)/\eps^2)$ against even an adaptive adversary. Furthermore, at each time $t$, the CDF estimate $\hat{F}_t$ returned by $\textsc{CdfEst}$ satisfies $\Exp (\hat{F}_t(i) - F_t(i))^2 \leq n/t$ for all $i \in [n]$.
\end{theorem}

\begin{proof}
For $\tau = 1, \dots, t$, let $y_\tau(i) = 1(x_\tau \leq i)$ denote the threshold function associated with $x_\tau$, and let $\hat{y}_\tau$ denote the estimate of $y_\tau$ defined by $\hat{y}_\tau(i) = n1(q_\tau = i)1(x_\tau \leq i)$, with $\Exp[\hat{y}_\tau \mid y_\tau] = y_\tau$. Observe that
\begin{equation*}
    \Var(\hat{y}_\tau(i) \mid x_\tau) = n^2\, 1(x_\tau \leq i) \Var(1(q_\tau = i)) = (n-1)\, 1(x_\tau \leq i).
\end{equation*}
We can now rewrite the algorithm's CDF estimate as $\hat{F}_t = \frac{1}{t}\sum_{\tau = 1}^t \hat{y}_\tau$, with $\Exp[\hat{F}_t \mid x_1, \dots, x_t] = \frac{1}{t}\sum_{\tau = 1}^t y_\tau = F_t$.
Since the estimate is correct in expectation, we have
\begin{equation*}
    \Exp(\hat{F}_t(i) - F_t(i))^2 = \frac{1}{t^2} \sum_{\tau=1}^t \Exp\left[\Var(\hat{y}_\tau(i) \mid x_\tau)\right] = \frac{n-1}{t} F_t(i) \leq \frac{n}{t}
\end{equation*}
for each $i \in [n]$. Furthermore, using Bernstein's inequality (or, in the case of an adaptive adversary, its martingale variant proved in \cite{freedman1975}), we obtain for any fixed $i \in [n]$ that
\begin{align*}
    \Pr\left[\left|\hat{F}_t(i) - F_t(i)\right| > \eps \right] \leq 2 \exp\left(-\frac{t^2\eps^2}{2(n-1)tF_t(i) + \frac{2}{3}nt\eps}\right) \leq 2 \exp\left(-\frac{t\eps^2}{3n}\right),
\end{align*}
which is less than $1/(4n)$ for $t \geq 3n \log(8n)/\eps^2$. Taking a union bound over $i$ gives the desired conclusion.
\end{proof}

\begin{remark}\label{remark:unbiased-estimation}
Assuming that queries are made uniformly at random, it is simple to verify that $\hat{y}_\tau$ is the unique unbiased estimate of $y_\tau$ computable from the feedback at time $\tau$.
\end{remark}

With this CDF estimate, the algorithm can extract a median estimate in the natural way, setting $\hat{m}_t = \min \{ i : \hat{F}_t(i) > 1/2 \}$. Noting that
\begin{equation*}
    E(\hat{m}_t) \leq \dist([\hat{F}_t(\hat{m}_t - 1), \hat{F}_t(\hat{m}_t)], 1/2) + \|\hat{F}_t - F_t\|_\infty = \|\hat{F}_t - F_t\|_\infty,
\end{equation*}
we obtain an immediate corollary.

\begin{corollary}\label{cor:median-est-upper-bound}
The median estimate $\hat{m}_t$ obtained from $\textsc{CdfEst}$ achieves anytime query complexity $O(n\log(n)/\eps^2)$ against even an adaptive adversary.
\end{corollary}

Given that these algorithms fail to incorporate the information feedback $1(x_t \leq q_t)$ reveals about other potential queries, the lower bounds which follow are somewhat surprising.

\subsection{Lower Bounds}\label{subsection:lower_bounds}

Although the performance of our algorithm for online median estimation is significantly worse than its stochastic counterpart, we show now that this degradation cannot be avoided in general. Our result builds upon a simpler lower bound for CDF estimation, matching the upper bound up to constant factors, by exploiting behavior that emerges in a regime of small $\eps$.

\begin{theorem}[Online CDF Estimation Lower Bound]\label{thm:cdf-est-lower-bound}
In the online comparison feedback model, no CDF estimation algorithm admits fixed-horizon query complexity $o(n\log(n)/\eps^2)$ against even a stochastic adversary.
\end{theorem}

\begin{proof}
For any $\eps \leq 1/(n+1)$, consider the family of distributions $\F_\eps$ indexed by $\sigma \in \{-1,+1\}^n$ admitting CDFs of the form
\begin{equation*}
    F_\sigma(i) = \frac{i}{n+1} + \sigma\eps, \quad \forall i \in [n], \qquad F_\sigma(n+1) = 1,
\end{equation*}
where the bound on $\eps$ guarantees that these CDFs are monotonic and hence well-defined. Next, we observe that simultaneously determining the biases of $n$ independent coins with probability at least $2/3$, where coin $i$ has bias $i/(n+1) \pm \eps$, requires $\Omega(n\log(n)/\eps^2)$ samples by standard KL divergence arguments. Hopefully, the direction of the reduction is now clear. 

Given such a testing setup, with unknown coin biases given by $\sigma \in \{-1,+1\}^n$, and an online CDF estimation algorithm $\A$, we can return a flip of coin $i$ as feedback whenever $\A$ queries $i$. By design, this feedback is indistinguishable from that given by the stochastic adversary which samples from the distribution with CDF $F_\sigma$. Thus, at a time horizon $T$, if we determine the bias of coin $i$ by rounding $\hat{F}_T(i)$ to the closest of $i/(n+1) \pm \eps$, all of our determinations are correct when $\|\hat{F}_T - F_\sigma\|_\infty < \eps$. If $\A$ has query complexity $T_0(n,\eps) = o(n\log(n)/\eps^2)$, then setting $T = \max\{T_0(n,\eps/3),15/\eps^2\} = o(n\log(n)/\eps^2)$ at the start and applying the Dvoretzky–Kiefer–Wolfowitz inequality (see Appendix Claim \ref{claim:DKW}) gives
\begin{equation*}
    \|\hat{F}_T - F_\sigma\|_\infty \leq \|\hat{F}_T - F_T\|_\infty + \|\hat{F}_T - F_\sigma\|_\infty \leq \eps/3 + \eps/3 < \eps
\end{equation*}
with probability at least $2/3$, violating the $\Omega(n\log(n)/\eps^2)$ testing lower bound.
\end{proof}

\begin{remark}
While this hardness result does not apply if we exclude the $\eps = O(1/n)$ regime, it can be extended slightly if we force the algorithm to learn $O(1/\eps)$ CDF values instead -- a more refined lower bound of $\Omega(\min\{n,1/\eps\}/\eps^2)$ is achievable with minor changes.
\end{remark}

Next, we construct a two-phase adversary which essentially forces any median estimation algorithm to perform full CDF estimation, implying a similar lower bound that matches the upper bound up to a $\log(n)$ factor. The precise details require some care, so we leave a full proof for the Appendix \ref{appendix:online}.

\begin{theorem}[Online Median Estimation Lower Bound]\label{thm:median-estimation-lower-bound}
In the online comparison feedback model, no median estimation algorithm admits fixed-horizon query complexity $o(n/\eps^2)$ against even an oblivious adversary.
\end{theorem}

\renewcommand{\proofname}{Proof Sketch}
\begin{proof}
At a high level, we show that online median estimation algorithms must simultaneously learn many quantiles (rather than just the median), which implies the strong lower bound. For a fixed time horizon $T$, we consider an oblivious adversary which returns samples from a distribution $\D$ for the first $T/2$ steps. This distribution is selected from a family such that learning $\eps$-good estimates for many of its quantiles from 0 to 1, in increments of $1/n$, mirrors the situation in \cref{thm:cdf-est-lower-bound} and requires $\Omega(n/\eps^2)$ samples. We will again require $\eps = O(1/n)$ so that this family is well-defined.

For the remaining $T/2$ steps, the adversary produces $k$ samples equal to $1$ and $T/2-k$ equal to $n$ for some random $k$ between 1 and $T/2$. Practically, this means that the algorithm's final median estimation error is actually its estimation error for a random quantile of the samples produced during the first phase. Hence, if the algorithm performs well, it has actually learned many of the said quantiles, and the previously mentioned hardness result kicks in. There are slight nuances expanded upon in the full proof which require that $T$ be sufficiently large (but still much smaller than $n/\eps^2$). However, because query complexity guarantees hold for all $T$ greater than some $T_0$, this does not present an obstacle to proving the hardness result.
\end{proof}
\renewcommand{\proofname}{Proof}

Since these lower bounds only pertain to the $\eps = O(1/n)$ regime, it is reasonable to verify that algorithms like \textsc{CdfEst} cannot perform better for larger $\eps$. A simple argument reveals that a lower bound of $\Omega(n)$ is unavoidable for all algorithms with the same querying behavior, for all $\eps$. 

\begin{claim}
In the online comparison feedback model, any median estimation algorithm which produces i.i.d.\ queries sampled uniformly from $[n]$ must have fixed-horizon query complexity $\Omega(n)$ against even a stochastic adversary.
\end{claim}
\begin{proof}
Consider an adversary which decides to always return $x_t = 1$ or $x_t = 2$ based on a fair coin flip before the rounds begin. Since the adversary's samples are constant, producing an $\eps$-good median (for any non-trivial $\eps$) is the same as correctly guessing 1 or 2. However, it is impossible for the algorithm to guess correctly with probability greater than $1/2$ unless it queries $q_t = 1$ for some $t$. The probability that the algorithm fails to query 1 after $T$ rounds is
\begin{equation*}
    \left(1 - \frac{1}{n}\right)^T \geq \exp\left(-T\frac{1/n}{1-1/n}\right) = \exp\left(-\frac{T}{n-1}\right),
\end{equation*}
since $1-x \geq \exp(-x/(1-x))$ for $x < 1$. Bounding this error probability below $1/4$ requires $T \geq T_0 = \log(4)(n-1)$.
\end{proof}

\subsection{A Lower Bound for Deterministic Algorithms}\label{section:deterministic}
We now show that any deterministic median estimation algorithm (and hence any such CDF estimation algorithm) incurs constant error in the non-stochastic setting, motivating our examination of randomized algorithms in the previous section.

\begin{proposition}\label{prop:deterministic-failure}
In the online comparison feedback model, no deterministic median estimation algorithm admits fixed-horizon error $o(1)$ against oblivious adversaries.
\end{proposition}

Fixing a deterministic median estimation algorithm $\A$, we will construct two oblivious adversaries, each represented as a sequence of elements in $[n]$, such that one forces $\A$ to admit constant error. For ease of exposition, these adversaries are presented as adaptive, but they can easily be converted into oblivious adversaries via simulation of $\A$, due to its deterministic nature. 
\newpage


\renewcommand{\proofname}{Proof Sketch}
\begin{proof}
\label{deterministic-intuition}
Suppose there are two adversarial sequences $L = l_1, \dots, l_T$ and $R = r_1, \dots, r_T$, such that the feedback $\A$ receives from $L$ is identical to the feedback it receives from $R$. That is, for all $t \in [T]$, $1(l_t \leq q_t) = 1(r_t \leq q_t)$. Because $\A$ cannot distinguish between the two, it must output the same median estimate $\hat{m}$ against both adversaries. With this in mind, we construct $L$ and $R$ so that no element is a $1/32$-good median estimate for both sequences, thereby forcing $\A$ to admit constant error against at least one sequence. For ease of exposition, we assume that $n$ is even.

To construct $L$ and $R$ with identical feedback, we must choose $l_t$ and $r_t$ at each step $t$ so that either both are at most $q_t$, or both are strictly greater than $q_t$. This is only feasible with knowledge of $q_t$; however, because $\A$ is deterministic, we can compute $q_t$ \textit{before} selecting $l_t$ and $r_t$ by simulating $\A$ on the prior history through time $t-1$. Having precomputed $q_t$, we consider the two sets $\{1, \dots, q_t \}$ and $\{q_t+1, \dots, n \}$, and, selecting the set with larger cardinality, we assign $l_t$ as its minimum element and $r_t$ as its maximum element. (In the case that $q_t = n/2$, we select $\{1, \dots, n/2 \}$.) Observe that any two elements belonging to the same set have identical feedback with respect to $q_t$, so $1(l_t \leq q_t) = 1(r_t \leq q_t)$. Additionally, since the cardinality of the larger set is always at least $n/2$, the support of $L$ lies entirely in $\{1, \dots, n/2 \}$, the ``left half'' of the support, and the support of $R$ lies entirely in $\{n/2, \dots, n \}$, the ``right half'' of the support. Thus, the only element that is in both the support of $L$ and the support of $R$ is $n/2$. Moreover, any estimate outside the support of a sequence incurs error $1/2$, so $\A$ must output $\hat{m} = n/2$ or suffer error $1/2$ for one of the two adversaries. 
We address the remaining issue of support overlap at $n/2$ in Appendix \ref{appendix:deterministic}.
\end{proof}
\renewcommand{\proofname}{Proof}

\section{Stochastic CDF Estimation}\label{section:stochastic}

The linear lower bounds in \cref{section:online} require that $\eps = O(1/n)$ in order for the distributions employed by the adversary to be well-defined.
An open question is whether similar linear lower bounds exist in the case that $\eps$ is constant. We examine CDF estimation for constant $\eps$ in the stochastic setting, where the samples $x_1, \dots, x_t$ are drawn i.i.d.\ from some fixed distribution $\mathcal{D}$ on $[n]$ with CDF $F$. Via simple reductions to noisy binary search, we show a logarithmic upper bound for CDF estimation in the stochastic setting. This suggests that an adversary such as the one presented in \cref{thm:cdf-est-lower-bound} cannot induce a linear lower bound for CDF estimation in the constant-$\eps$ regime; the lower bound may truly be sub-linear, or perhaps a more adaptive adversary is required.

For ease of exposition, we adjust our benchmark in the stochastic setting to define error with respect to the population CDF $F$, instead of the empirical CDF as in \cref{section:online}. A CDF estimate $\hat{F}$ is evaluated with respect to the error
$$E(\hat{F}) = \| \hat{F} - F\|_\infty.$$

A key component in our algorithm for stochastic CDF estimation is stochastic quantile estimation, which reduces to the noisy binary search problem studied by ~\cite{karp2007noisy}, rephrased below.

\begin{theorem}[Noisy Binary Search, rephrased ~\cite{karp2007noisy}] \label{thm:noisy_bsearch}
Suppose there are $n$ Bernoulli random variables 
with unknown, non-decreasing means $p_1 \leq \dots \leq p_n$. Additionally, define $p_0 = 0$ and $p_{n+1} = 1$. At each time step $t$, an algorithm chooses an index $i \in [n]$ and observes $y_t \sim \Ber(p_i)$. For any $\tau \in [0,1]$, there is an algorithm that makes $O(\log(n))$ queries in expectation and outputs an estimate $m$ such that $\dist([p_m, p_{m+1}], \tau) \leq 1/8$ with probability at least $3/4$.
\end{theorem}

Combining noisy binary search with a simple confidence boosting procedure, we can estimate any $\tau$-quantile with a logarithmic number of queries.

\begin{lemma}\label{lem:boost_stochastic_median} For any $\tau \in [0, 1]$, there is an algorithm $\A(\tau)$ which makes at most $O(\log(n))$ queries in the stochastic comparison feedback model and outputs an estimate $w_\tau \in [n]$ such that
$$\dist([F(w_\tau-1), F(w_\tau)], \tau]) \leq 1/8$$
with probability at least $.99$.
\end{lemma}
\begin{proof}
As described in \cref{section:intro}, stochastic median estimation reduces to the noisy binary search problem, because $A_q = 1(x \leq q)$ is a Bernoulli random variable with mean $\Exp[A_q] = F(q)$. So querying an element $q$ is equivalent to flipping a coin with bias $F(q)$, and since the CDF is a monotone increasing function, $F(1) \leq \dots \leq F(n)$. While the $O(\log(n))$ bound on the number of queries made by the algorithm in \cref{thm:noisy_bsearch} is only in expectation, we can convert this algorithm into a deterministic algorithm with the same probabilistic guarantees, by running it for a constant number of trials and halting each trial before $O(\log(n))$ steps. By a median-of-medians procedure similar to Lemma \ref{lemma:confidence-boosting}, we can boost the confidence of the resulting algorithm to achieve success with probability at least $.99$ while paying only a constant multiplicative factor in the number of queries.
\end{proof}

We are now equipped to present our CDF estimation procedure.
For each $\tau \in \{ 1/8, 2/8, \dots, 1\}$, the algorithm $\textsc{StochasticCDF}$ computes the quantile estimate $w_\tau = \A(\tau)$, where $\A(\tau)$ is the algorithm from Lemma \ref{lem:boost_stochastic_median}, and then for all $j \in [n]$, it assigns 
\begin{align}
\hat{F}(j) = \max \{\tau: w_\tau \leq j \}.
\end{align}

\begin{theorem}\label{thm:stochastic_cdf}
With $O(\log(n))$ queries, $\textsc{StochasticCDF}$ produces a CDF estimate $\hat{F}$ such that $E(\hat{F}) \leq 1/4$ with probability at least $3/4$.
\end{theorem}




\begin{proof}
Condition on the event that $w_\tau$ is an $1/8$-good estimate for each $\tau \in \{ 1/8, 2/8 \dots, 1 \}$ simultaneously. Since each individual estimate is $1/8$-good with probability at least $.99$, this event occurs with probability at least $3/4$. Then for all $\tau$, $\dist([F(w_\tau-1), F(w_\tau)], \tau]) \leq 1/8.$ Therefore, $F(w_\tau - 1) \leq \tau + 1/8$ and $F(w_\tau) \geq \tau - 1/8$. 

Next consider the assignment of $\hat{F}(j)$. For any $j \in [n]$, if $\hat{F}(j) = \tau < 1$, then by definition, $w_\tau \leq j$. Additionally, since $\tau$ is the largest index such that $w_\tau \leq j$, we have $j < w_{\tau + 1/8}$. Therefore, $F(j) \leq F(w_{\tau + 1/8} - 1) \leq \tau + 1/4$ and $F(j) \geq F(w_\tau) \geq \tau - 1/8$. Thus $ \tau - 1/8 \leq F(j) \leq \tau + 1/4$, so $|\hat{F}(j) - F(j)| \leq 1/4$. If $\hat{F}(j) = \tau = 1$, then $w_1 \leq j$, so $1 \geq F(j) \geq F(w_1) \geq 1 - 1/8$, and $|\hat{F}(j) - F(j)| \leq 1/4$.
 
Each quantile estimate requires a call to $\A(\tau)$, which, by Lemma \ref{lem:boost_stochastic_median}, takes $O(\log(n))$ time, so computing the quantiles $w_{1/8}, w_{2/8}, \dots, w_1$ requires at most $O(\log(n))$ queries in total.
\end{proof}

\section{Mean Estimation}\label{section:means}

Next, we examine the simpler problem of mean estimation. Our algorithm is similar to the CDF case, querying uniformly at random to collect unbiased estimates and returning their average. This procedure works against even adaptive adversaries while still achieving near-optimal performance across the board.

Formally, our mean estimation algorithm $\textsc{MeanEst}$ generates i.i.d.\ queries $q_1, q_2, \dots$ sampled uniformly from $[n]$, and maintains mean estimate
\begin{equation}
    \hat{\mu}_t = 1 + \frac{n}{t} \sum_{\tau=1}^t 1(x_\tau > q_\tau).
\end{equation}
Recall that our error is measured as $E(T) = |\hat{\mu}_t - \mu_t|/n$, where $\mu_t$ is the empirical mean of $x_1, \dots, x_t$.

\begin{theorem}[Online Mean Estimation Upper Bound]\label{thm:online-means-ub}
In the online comparison feedback model, the mean estimate $\hat{\mu}_t$ returned by $\textsc{MeanEst}$ incurs mean squared error $\Exp[(\hat{\mu}_t - \mu_t)^2/n^2] \leq 1/(4t)$ at each time $t$. Hence, $\textsc{MeanEst}$ has anytime query complexity $1/\eps^2$ against even an adaptive adversary.
\end{theorem}

\begin{proof}
The following observation motivates our choice of $\hat{\mu}_t$:
\begin{equation*}
    \mu_t = \frac{1}{t}\sum_{\tau=1}^t x_\tau = 1 + \frac{1}{t}\sum_{\tau=1}^t \Pr(x_\tau > q_\tau).
\end{equation*}
Indeed, we can view $\hat{\mu}_t$ as the average of $t$ estimates of the form $\hat{x}_\tau = 1 + n 1(x_\tau > q_\tau)$, where $\Exp[\hat{x}_\tau \mid x_\tau] = x_\tau$ and $\Var(\hat{x}_\tau \mid x_\tau) \leq n^2/4$. Consequently, the mean squared error $\Exp[(\hat{\mu}_t - \mu_t)^2/n^2]$ is at most $1/(4t)$. Markov's inequality then gives that $|\hat{\mu}_t - \mu|/n \leq \sqrt{1/t}$ with probability at least $3/4$, implying the desired query complexity.
\end{proof}

\begin{remark}
Similarly to Remark \ref{remark:unbiased-estimation}, if we assume that queries are made uniformly at random, one can verify that $\hat{x}_\tau$ has minimum variance among all unbiased estimates of $x_\tau$ computable from the feedback at time $\tau$.
\end{remark}

Next, we prove that the mean squared error guarantee of \textsc{MeanEst} is tight up to constant factors, via a lower bound on conditional variance.

\begin{proposition}[Online Mean Estimation Lower Bound]
In the online comparison feedback model, there exists a stochastic adversary which forces even fixed-horizon mean estimation algorithms to admit
mean squared error $\Exp[(\hat{\mu}_T - \mu_T)^2/n^2] = \Omega(1/T)$.
\end{proposition}

\begin{proof}
Consider the adversary which returns i.i.d.\ samples $x_t \sim \Unif([n])$ for $t \in [T]$, and let $b_t = 1(x_t \leq q_t)$ denote the feedback revealed to the algorithm during round $t$. Next, consider the event $A_t$ that $x_t$ lies within the larger of two intervals determined by $q_t$, i.e., $\# \{ x \in [n] : 1(x \leq q_t) = b_t \} \geq n/2$. Now, the variance of a random variable distributed uniformly on the integers within an interval of size at least $n/2$ is at least $(n^2-4)/48$, so $\Var(x_t \mid b_t, A_t) = \Omega(n^2)$ for all $t \in [T]$. Noting that each event $A_t$ occurs with probability at least 1/2, the law of total variance and independence of samples give that
\begin{equation*}
    \Var(\mu_T \mid b_1, \dots, b_T) = \frac{1}{T^2} \sum_{t=1}^T \Var(x_t \mid b_t) \geq \frac{2}{T^2} \sum_{t=1}^T \Var(x_t \mid b_t, A_t) = \Omega(n^2/T).
\end{equation*}
Finally, the law of total expectation and the bias-variance decomposition for estimators imply
\begin{equation*}
    \Exp\bigl[ (\hat{\mu}_T - \mu_T)^2/n^2 \bigr] \geq \Exp\bigl[\Var(\overline{x}_T \mid b_1, \dots, b_T)\bigr]/n^2 = \Omega(1/T),
\end{equation*}
as desired.
\end{proof}

As a final note, we observe that the choice of empirical versus distributional benchmark is not a triviality when analyzing online mean estimation against a stochastic adversary. Indeed, we can easily obtain a stronger $\Omega(1/\eps^2)$ query complexity lower bound against a distributional benchmark by noting that distinguishing between the two distributions which place equal mass on $1$ and $n$ except for a $\pm \eps n$ bias requires $\Omega(1/\eps^2)$ queries. (This holds even in the full-feedback setting, by standard KL arguments.) The issue in this setting is that the empirical mean is only $\eps$-close to the true mean after $\Omega(1/\eps^2)$ samples, which is the same magnitude as our desired lower bound.

\section{Generalizations and Future Work}\label{section:generalizations}

An immediate extension of these results is to the continuous setting, where queries and samples are selected from $[0,1]$. In this case, we need to specify some resolution $\delta$ of interest and can examine the complexity of maintaining a value within distance $\delta$ of an $\eps$-good estimate. However, by choosing $\delta = \Omega(1/n)$, it is straightforward to transition between this setting and ours, so we opted to present discrete results for ease of exposition. More interestingly, there are several high-dimensional analogs of our online estimation problem.

\subsection{Graph Medians, Geometric Medians, and Online Convex Programming}

We start with a discrete example, fixing some graph $G$ on $n$ nodes. In this case, we can imagine that each sample $x_t$ and query $q_t$ are nodes of $G$, where an edge leaving $q_t$ along a shortest path to $x_t$ is given as feedback, mirroring the setup for graph binary search. The natural analog of an empirical median is a minimizer of the potential $\Phi(m) = \sum_t d(x_t, m)$, where $d$ is the shortest path distance on $G$. When samples are distributed uniformly over the vertex set, this corresponds to the standard notion of a graph median, as examined in \cite{dereniowski2020}. Our results apply directly to this setting when $G$ is taken to be a path. 

Similarly, we can extend our sample and query space to $[0,1]^d$, where a unit vector in the direction $x_t - q_t$ is given as feedback. The natural object of estimation is now a minimizer of the potential $\Phi(m) = \sum_t \| x_t - m \|$, commonly known as the geometric median. Generalizing further, we can connect this to the following online convex programming problem. At each time $t$, the adversary produces a convex function $f_t:[0,1]^d \to [0,1]$, the algorithm queries a point $q_t \in [0,1]^d$, and a subgradient $g_t \in \partial f_t(q_t)$ is given as feedback. We observe that this matches the previous setting upon taking $f_t(q) = \|x_t - q\|$. In this case, we hope to maintain an estimate $\hat{m}$ such that the function $F(m) = \sum_t f_t(m)$ has a small subgradient within the subdifferential set $\partial F(\hat{m})$. To make this connection concrete, we note that upon taking $d=1$ and $f_t(q) = |x_t - q|$, the estimation error of a median estimate $\hat{m}$ for $x_1, \dots, x_T$ can be expressed as
\begin{equation*}
    E(\hat{m}) = \dist\left(\left[F_T(\hat{m}-1), F_T(\hat{m})\right],\frac{1}{2}\right) = \frac{1}{2T} \min_{g \in \partial \sum_{t=1}^T f_t(\hat{m})} |g|.
\end{equation*}

\subsection{Intermediate Adversaries}

In these generalized frameworks, we expect that algorithms which maintain good estimates in the non-stochastic adversarial setting, across parameter regimes, will also exhibit the uniform querying behavior that appeared in this work. Thus, as an avenue for future research, we propose analyzing these online estimation problems against intermediate adversaries which sample from nearly stationary distributions, i.e., with some bound on the statistical distance between the distributions at time $t$ and $t+1$.

\acks{Sloan Nietert is supported by the National Science Foundation Graduate Research Fellowship under Grant DGE-1650441. Michela Meister is supported by the Department of Defense (DoD) through the National Defense Science \& Engineering Graduate (NDSEG) Fellowship Program.We are grateful to Bobby Kleinberg for several fruitful conversations, which, in particular, introduced us to the online comparison feedback model and the convex optimization perspective. We would also like to thank Spencer Peters for several useful adversary and algorithm suggestions, as well as Nika Haghtalab and Abhishek Shetty for various pieces of feedback and advice.}

\bibliography{references}

\appendix

\section{Preliminaries}\label{appendix:preliminaries}
\renewcommand{\proofname}{Proof of Lemma \ref{lemma:median-to-quantile}}
\begin{proof}
Fix such an algorithm $\A$ and suppose that $\tau > 1/2$. Then, consider the modified algorithm $\A'$ which simulates $\A$ but replaces each bit of feedback of the form $1(x_t \leq q_t)$ with $1(x_t \leq q_t)B_t$, where $B_t \sim \Ber(1/(2\tau))$ are iid. In the stochastic case, if $x_t \sim \D$ and $1(x_t \leq q) \sim \Ber(\alpha_q)$ for $\alpha_q = \Pr_{X \sim \D}[X \leq q]$, then the modified feedback has distribution $\Ber(\alpha_q/(2\tau))$. Searching for $m$ such that $\alpha_m/(2\tau) = 1/2$ corresponds to finding $m$ such that $\Pr[X \leq m] = \tau$, as desired, with error scaled by a factor of $2\tau \in (1, 2]$.

For the non-stochastic case, the desired conclusion follows upon noting that the modified feedback is equal to $1(x_t' \leq q_t)$ for $x_t' = x_t B_t + (n+1)(1-B_t)$. Indeed, an empirical median of $x_1', \dots, x_T'$ is an approximate $\tau$-quantile for $x_1, \dots, x_T$, with error scaled by the same $2\tau$ factor, so long as $\sum_{t=1}^T 1(x'_t \leq q_t)$ does not deviate too much from $\sum_{t=1}^T 1(x_t \leq q_t)/2\tau$. A query complexity lower bound of $\Omega(1/\eps^2)$ ensures that any deviations are sufficiently small via an application of the Azuma-Hoeffding inequality for martingale increments. Of course, it is essential here that each sample $x_t$ produced by the adversary is independent of the coin flip $B_t$. The case of $\tau < 1/2$ follows in the same way with modified feedback $1(x_t \leq q_t)B_t + (1 - B_t)$, where $B_t \sim \Ber(1/(2(1-\tau)))$ are again iid.
\end{proof}
\renewcommand{\proofname}{Proof}

\renewcommand{\proofname}{Proof of Lemma \ref{lemma:fixed-horizon-reduction}}
\begin{proof}
Consider any online median or CDF estimation algorithm $\A$. By the definition of $\mathcal{B}$, for some $T_0$, there exists a sequence $x_1, \dots, x_{T_0}$ forcing $\A$ to admit error $E(T_0) \geq 1/16$, so in the same way, for $T_1 = 32T_0$, there exists a sequence $x_{T_0 + 1}, \dots, x_{T_0 + T_1}$ forcing $\A$ to admit error at least $1/16$ on that sequence and error $E(T_0+T_1) \geq 1/16 - T_0/(T_0 + T_1) \geq 1/32$ overall on the sequence $x_1, \dots, x_{T_0+T_1}$. 

Now consider any online mean estimation algorithm $\A$. By the same argument, for some $T_0$, there exists a sequence $x_1, \dots, x_{T_0}$ forcing $\A$ to admit error $E(T_0) \geq 1/16$, and for $T_1 = 32T_0$, there exists a sequence $x_{T_0 + 1}, \dots, x_{T_0 + T_1}$ forcing $\A$ to admit error at least $1/16$ on that sequence and error $E(T_0+T_1) \geq 1/16 \cdot (T_1/(T_0 + T_1)) - T_0/(T_0 + T_1) \geq 1/16 \cdot 32/33 - 1/33 > 1/32$ overall on the sequence $x_1, \dots, x_{T_0+T_1}$.

In both of the above cases, we can repeat the process ad infinitimum to produce a sequence $x_1, x_2, \dots$ such that $E(\sum_{j=0}^k T_032^j) \geq 1/32$ for all $k \in \NN$, concluding the proof. 
\end{proof}
\renewcommand{\proofname}{Proof}

\section{Online Median Estimation Lower Bound}\label{appendix:online}

Our lower bound for online median estimation is built above several claims. First, we simply state the well-known Dvoretzky–Kiefer–Wolfowitz inequality.

\begin{claim}\label{claim:DKW}
If $x_1, \dots, x_T$ are iid samples from a distribution $\D$ on $[n]$ with cdf $F$, then
\begin{equation*}
    \norm{F_T - F}_\infty = \max_i |F_T(i) - F(i)| \leq \sqrt{\ln(2/\delta)/2T} = O\left(\sqrt{\log(1/\delta)/T}\right)
\end{equation*}
with probability $1 - \delta$.
\end{claim}

Next, we present an initial information-theoretic lower bound.

\begin{claim}\label{claim:quant-lb-1}
Fix $n = 4k$ for some positive, even integer $k$, and fix $\eps \leq 1/(2n)$. Suppose that $\A$ is an algorithm in the comparison feedback model which makes $T$ queries against a stochastic adversary and learns estimates for quantiles $(k+1)/n, (k+3)/n, \dots (3k-1)/n$ of its distribution such that at least $2/3$ are $\eps$-good with probability greater than $3/4$. Then $T = \Omega(n/\eps^2)$.
\end{claim}

\begin{proof}
Assume for ease of exposition that $n = 4k$ for some positive, even integer $k$. We define $\mathcal{F}_\eps$ to be the family of distributions indexed by $\sigma \in \{-1,+1\}^{k}$ with CDFs of the form
\begin{equation*}
    F_\sigma(i) = \begin{cases}
        \frac{i}{n} & i \text{ even, } i \leq k, \text{ or } i \geq 3k,\\
        \frac{i}{n} + \sigma_{(i-k+1)/2}\, \alpha_i \,\eps & \text{otherwise},
    \end{cases}
\end{equation*}
where $\alpha_i = 2 - 4\left|\frac{i}{n} - \frac{1}{2}\right| \in [1,2]$ for $k < i < 3k$. That is, $F_\sigma$ corresponds to a distribution which, for $j = 1,\dots,k$, places mass $1/n + \sigma_j \Theta(\eps)$ on element $k + 2j - 1$ and mass $1/n - \sigma_j \Theta(\eps)$ on element $k + 2j$, while placing uniform mass $1/n$ on the remaining elements. We require that $\eps \leq 1/(2n)$ so that all of these masses are positive.

First, we observe that testing whether $k$ independent coins have bias $1/2 + 2\eps$ or bias $1/2 - 2\eps$, each with probability strictly greater than $1/2$, requires $\Omega(n/\eps^2)$ samples by standard KL divergence arguments. If $\A$ performs as claimed against all distributions in $\mathcal{F}_\eps$, we will see that it can perform this testing procedure with $T$ samples, implying the desired lower bound. Indeed, given a set of such coins with the signs of their biases given by an unknown $\sigma \in \{-1,+1\}^k$, sample a random permutation $\pi \in S_k$ uniformly at random. Then, if $\A$ queries an element $i = k + 2j - 1$ in the region of interest, use coin $\pi(j)$ and the procedure outlined in proof of Lemma \ref{lemma:median-to-quantile} to return the flip of a coin with bias $i/n + \sigma_{\pi(j)} \alpha_i \eps$. Importantly, this can be done without knowledge of $\sigma_j$, and the $\alpha_i$ values were chosen precisely with this procedure in mind. Otherwise, if $\A$ queries $i$ which is even or outside of the region of interest, return the flip of a coin with bias $i/n$ as feedback. By design, this feedback is indistinguishable to the algorithm from that which it would receive in the offline comparison feedback setting against the distribution with CDF $F_\sigma$.

Now, to determine the bias of coin $j$, we examine the algorithm's estimate $m_\tau$ for quantile $\tau = (k + 2\pi^{-1}(j) - 1)/n$. If $m_\tau$ is $\eps$-good, then one can show that $m_\tau = k + 2\pi^{-1}(j) - 1 + (1 - \sigma_j)/2$, from which we can extract the value of $\sigma_j$. By the algorithm's estimation guarantee, at least proportion $1/2$ of the quantiles of interest are $\eps$-good with constant probability. Since $\pi$ was sampled uniformly at random, this implies that each of these bias determinations is correct with probability greater than $2/3 \cdot 3/4 = 1/2$. Hence, the KL lower bound kicks in and requires $T = \Omega(n/\eps^2)$.
\end{proof}

\begin{claim}\label{claim:random-index}
Let $B_1, \dots, B_m$ be random events and take $U \sim \Unif([m])$ to be an independently selected index. If $\Pr[B_U] \leq 1/15$, then the probability that more than $1/3$ of the events occur is less than $1/5$.
\end{claim}
\begin{proof}
Let $\alpha$ be the random variable denoting the fraction of events which occur. Then,
\begin{equation*}
    \frac{1}{15} \geq \Pr\left[B_U\right] \geq \Pr\left[B_U \text{ and } \alpha > \frac{1}{3}\right] = \Pr\left[B_U \bigg| \alpha > \frac{1}{3} \right] \Pr\left[\alpha > \frac{1}{3}\right] > \frac{1}{3} \Pr\left[\alpha > \frac{1}{3}\right],
\end{equation*}
implying the desired result.
\end{proof}

\begin{claim}\label{claim:quant-lb-2}
Fix $n = 4k$ for some positive, even integer $k$, and fix $\eps \leq 1/(2n)$. Suppose that $\A$ is an algorithm in the online comparison feedback model which, for any fixed time horizon $T \geq T_0$, produces estimates for quantiles $(k+1)/n, (k+3)/n, \dots (3k-1)/n$ of $F_T$, such that an estimate selected uniformly at random is $\eps$-good with probability at least $14/15$. Then, $T_0 = \Omega(n/\eps^2)$.
\end{claim}

\begin{proof}
If the algorithm performs as stated, then Claim \ref{claim:random-index} implies that at least proportion $2/3$ of the estimates are $\eps$-good with probability at least $4/5$, whenever $T \geq T_0$. Further, we know by \ref{claim:DKW} that $\|F - F_T\|_\infty \leq \eps$ with probability at least $24/25$ whenever $T \geq 3/\eps^2$. Hence, if $T \geq \max\{T_0,2/\eps^2\}$, we have that at least $2/3$ of the estimates are $(2\eps)$-good for $F$ with probability at least $19/25 > 3/4$. Thus Claim \ref{claim:quant-lb-1} forces $T_0 + 2/\eps^2 \geq \max\{T_0,2/\eps^2\} = \Omega(n/\eps^2)$, implying the desired lower bound.
\end{proof}

At last, we are prepared to give a full proof of the theorem.

\renewcommand{\proofname}{Proof of Theorem \ref{thm:median-estimation-lower-bound}}
\begin{proof}
Suppose that a fixed-horizon online median estimation algorithm $\A$ admits error
\begin{equation*}
    E(T) = \dist\left([F_T(\hat{m}_T-1),F_T(\hat{m}_T)], \frac{1}{2}\right) \leq f(T,n)
\end{equation*}
with probability at least $14/15$, for any time horizon $T$ and support size $n$. Here we have increased the query complexity success probability from $3/4$ to $14/15$ to allow for a cleaner presentation, though this is not necessary. We will essentially show that $f(T,n) = \Omega(\sqrt{n/T})$ for all $n \geq n_0 = O(1)$ and $T \geq T_0 = \Omega(n^3)$, proving the theorem. For convenience, we restrict the parameters a bit further (without sacrificing the conclusion), requiring that $n = 4k$ for some positive integer $k \geq 3$ and that $T = 2nm$ for some positive integer $m \geq 8n^2$. Using a carefully selected class of adversaries, we will reduce this instance of median estimation to the quantile estimation of Claim \ref{claim:quant-lb-2} and obtain our lower bound. This conclusion already holds if $f(T,n) > 1/(4n)$, so we will assume that the algorithm has good performance, with $f(T,n) \leq 1/(4n)$.

For any distribution $\D$ on $[n]$ with cdf $F$, consider the following oblivious adversary, which starts by selecting an integer $j$ from the set $\{k+1, k+3, \dots 3k-1\}$ uniformly at random. For times $1, \dots, T/2$, this opponent returns independent samples from $\D$. For times $T/2+1, \dots, T/2+jm$, the adversary returns $n$, and for times $T/2+jm+1, \dots, T$, the adversary returns $1$. With these choices, the empirical cdf $F_T$ is given by
\begin{equation*}
    F_T(i) = \frac{1}{2}F_{\frac{T}{2}}(i) + \left(\frac{1}{2} - \frac{j}{2n}\right) + \frac{j}{2n}1(i = n).
\end{equation*}
Thus, if $\hat{m}_T < n$, we have
\begin{align*}
    E(T) &= \dist\left(\left[\frac{1}{2}F_{\frac{T}{2}}(\hat{m}_T-1) + \frac{1}{2} - \frac{j}{2n}, \frac{1}{2}F_{\frac{T}{2}}(\hat{m}_T) + \frac{1}{2} - \frac{j}{2n}\right], \frac{1}{2}\right)\\
    &= \frac{1}{2} \dist\left(\left[F_{\frac{T}{2}}(\hat{m}_T-1), F_{\frac{T}{2}}(\hat{m}_T)\right], \frac{j}{n} \right).
\end{align*}
This is nearing the desired quantile estimation problem. The case of $\hat{m}_T = n$ is less pleasant but can be avoided since we are considering a reasonable algorithm. To see this, we first note that our lower bound on $T$ implies $\|F_{T/2}-F\|_\infty \leq 1/n$ with probability at least $3/4$. Furthermore, we take into account that the relevant distributions $\D$ will all be selected from the the family $\mathcal{F}_\eps$ introduced in the previous claim, for some $\eps \leq 1/(2n)$, so $F(n-1) = \frac{n-1}{n}$. Lastly, $j \leq 3n/4 \leq n - 3$ by our choice of $n$. With these bounds, we can rule out the possibility that $\hat{m}_T = n$, since this implies
\begin{align*}
   1/(4n) \geq f(T,n) \geq E(T) &= \dist\left(\left[\frac{1}{2}F_{\frac{T}{2}}(n-1) + \frac{1}{2} - \frac{j}{2n}, \frac{1}{2}F_{\frac{T}{2}}(n) + \frac{1}{2}\right], \frac{1}{2}\right)\\
    &= \frac{1}{2} \max\left\{F_{\frac{T}{2}}(n-1) - \frac{j}{n}, 0\right\}\\
    &\geq \frac{1}{2} \max\left\{\frac{n-1}{n} - \frac{n-3}{n}, 0\right\} - 1/n = \frac{1}{n}
\end{align*}
with positive probability, a contradiction. Hence, $\hat{m}_T < n$, and we have
\begin{equation*}
    \dist\left(\left[F_{T/2}(\hat{m}_T-1), F_{T/2}(\hat{m}_T)\right], \frac{j}{n} \right) \leq 2f(T,n) \leq 1/(2n)
\end{equation*}
with probability at least $14/15$. That is, our algorithm has the quantile estimation power assumed in Claim \ref{claim:quant-lb-2} (at least for the relevant family of distributions) for the choice of $\eps = 2f(T,n)$, so it follows that $f(T,\delta,n) = \Omega(\sqrt{n/T})$, as desired.
\end{proof}
\renewcommand{\proofname}{Proof}

\section{Lower Bound for Deterministic Algorithms}\label{appendix:deterministic}
\renewcommand{\proofname}{Proof of Proposition \ref{prop:deterministic-failure}}
\begin{proof}
Following the intuition of ~\cref{deterministic-intuition}, for a fixed algorithm $\A$, we wish to construct two sequences $L = l_1, \dots, l_T$ and $R = r_1, \dots, r_T$ with identical feedback such that no element is a $1/32$-good median estimate of both sequences. The construction presented in ~\cref{section:online} is almost sufficient, however it fails against an algorithm that makes an equal number of queries to $n/2$ as to $n/2 + 1$; in this case, $n/2$ is the median estimate of both sequences. To address this, we construct two-phase adversaries, echoing the strategy from \cref{thm:cdf-est-lower-bound}. For the first $T/2$ time steps, $l_t$ and $r_t$ are assigned exactly as in ~\cref{deterministic-intuition}. After time $T/2$, $L$ and $R$ output the same sequence $w = w_1, \dots, w_{T/2}$, made up of 1's and $n$'s, which artificially ``shifts'' the medians of $L$ and $R$ in the case that $n/2$ is a good median for both sequences in the first phase.

For ease of presentation, let $n$ be even, and let $T$ be divisible by 16. To construct $L$ and $R$, for all $t \in [ T/2]$, consider the two sets, $\{1, \dots, q_t \}$ and $\{q_t+1, \dots, n \}$, and, fixing the set with larger cardinality, assign $l_t$ the minimum element and $r_t$ the maximum element. Observe that any two elements belonging to the same set have identical feedback with respect to $q_t$, so $1(l_t \leq q_t) = 1(r_t \leq q_t)$. To choose $w$, 
let $p = (T/2)\sum_{t=1}^{T/2} 1(1 \leq q_t \leq n/2-1)$; if $|1/2 - p| > 1/8$, let $w$ be a sequence of $T/4$ 1's followed by $T/4$ $n$'s, and if $|1/2-p| \leq 1/8$, let $w$ be a sequence of $T/8$ 1's followed by $3T/8$ $n$'s. Then for all $T/2 +1 \leq t \leq T$, set $l_t = r_t = w_{t - T/2}$.

We now examine two cases --- the first where $p$ is far from $1/2$ and the second where $p$ is close to $1/2$ --- and show that in both situations $\A$ incurs error at least $1/16$ against at least one of $L$ or $R$. 

First consider the case that $|1/2 - p| > 1/8$. In this case, $w$ contains an equal number of 1's and $n$'s, so for both sequences $L$ and $R$ the median of the first $T/2$ elements is equal to the median of the entire sequence. Observe that $l_1, \dots, l_{T/2}$ is supported on $\{ j \in [n]: 1 \leq j \leq n/2 \}$ and $r_1, \dots, r_{T/2}$ is supported on $\{ j \in [n]: n/2 \leq j \leq n \}$, so the only elements of $L$ that are at least $n/2 +1$ are the $T/4$ $n$'s from $w$, and the only elements of $R$ that are at most $n/2 -1$ are the $T/4$ 1's from $w$. As in case 1, because the feedback from $L$ and $R$ is indistinguishable to $\A$, $\A$ must output the same estimate $\hat{m}$ against both adversaries. If $\hat{m} < n/2$, then $\A$ incurs error at least $1/4$ with respect to $R$, and if $\hat{m} > n/2$, $\A$ incurs error at least $1/4$ with respect to $L$. Therefore $\A$ must output $\hat{m} = n/2$.

However, by the bound on $p$ for this case, we see that $n/2$ cannot be a $1/16$-good median for both $L$ and $R$. By the definitions of $p$, $R$, and $L$, at least $pT/2$ of the elements in $r_1, \dots, r_{T/2}$ are $n$, and at least $(1-p)T/2$ of the elements in $l_1, \dots, l_{T/2}$ are $1$. Since $|1/2 - p| > 1/8$, either $p > 5/8$ or $p < 3/8$. If $p > 5/8$, then strictly more than $5T/16$ of the elements in $r_1, \dots, r_{T/2}$ are $n$, and since $w$ contains $T/4$ $n$'s, $R$ contains strictly more than $9T/16$ $n$'s in total, so the median of $R$ is $n$, and any $\hat{m} \neq n$ is not a $1/16$-good median for $R$. If $p < 3/8$, then strictly more than $5T/16$ of the elements in $l_1, \dots, l_{T/2}$ are $1$, and since $w$ contains $T/4$ 1's, $L$ contains strictly more than $9T/16$ 1's in total, so 1 is the median of $L$ and any $\hat{m} \neq 1$ is not a $1/16$-good median for $L$. So $\A$ incurs error at least $1/16$ for one of $L$ or $R$ if $|p - 1/2| > 1/8$.

Next consider the case that $|1/2 - p| \leq 1/8$. In this case, $w$ is a string with $T/8$ 1's followed by $3T/8$ $n$'s. For every $t \in [T/2]$ such that $1 \leq q_t \leq n/2 - 1$, $r_t = n$, so $n$ occurs $pT/2$ times in the sequence $r_1, \dots, r_{T/2}$, and by the setting of $w$, $n$ occurs $3T/8$ times in the sequence $r_{T/2+1}, \dots, r_T$. Since $p \geq 3/8$, $n$ occurs at least $9T/16$ times throughout the entire sequence $r_1, \dots, r_T$, and therefore $n$ is the median of $R$. Moreover, because $9/16 \geq 1/2 + 1/16$, any estimate $\hat{m} \neq n$ incurs error at least $1/16$ against $R$.

Next we show that $n$ is not an $1/16$-good median estimate for $L$. Let $F_L$ be the CDF of the sequence $L$, and observe that $l_t \neq n$ for all $t \in [T/2]$. Thus, the only $n$'s that occur in $L$ are the $3T/8$ $n$'s in the second half of the sequence. Therefore, $F_L(n-1) = 10/16$, so $[F_L(n-1), F_L(n)] \cap [7/16, 9/16] = \emptyset$, and thus $n$ is not a $1/16$-good median estimate for $L$. Because the feedback from $L$ and $R$ is indistinguishable to $\A$, $\A$ must output the same estimate $\hat{m}$ against both sequences. However, if $\hat{m} \neq n$, $\A$ incurs error at least $1/16$ with respect to $R$, and if $\hat{m}=n$, $\A$ incurs error at least $1/16$ with respect to $L$. So $\A$ incurs error at least $1/16$ against either $L$ or $R$ if $|p - 1/2| \leq 1/8$.

Since in both cases $\A$ incurs error at least $1/16$ against either $L$ or $R$, there is no deterministic median estimation algorithm admitting error $E(T) < 1/16$.
\end{proof}

\section{Confidence Boosting}\label{appendix:confidence-boosting}

Finally, we motivate our restriction to constant confidence probability $3/4$ in the definition of query complexity, at least for the purposes of upper bounds. Note that the algorithms we provide in Sections \ref{section:online}
and \ref{section:means} have query complexities which are well-suited for the somewhat awkward maxima present in this statement.

\begin{lemma}[Confidence Boosting]\label{lemma:confidence-boosting}
Let $\A$ be an online median, CDF, or mean estimation algorithm with fixed-horizon or anytime query complexity $T_0(n,\eps)$ against some class of adversaries in the comparison feedback model. Then, for any $\delta > 0$, there exists a modified algorithm $\A'$ admitting query complexity $O(\max\{T_0(\eps/2) \Theta(\log(1/\delta)),\Theta(n \operatorname{polylog}(1/\delta)/\eps^2)\})$ (in the median/CDF case) or $O(\max\{T_0(\eps/2) \Theta(\log(1/\delta)),\Theta(\operatorname{polylog}(1/\delta)/\eps^2)\})$ (in the mean case) with confidence probability $1 - \delta$ in the same setting.
\end{lemma}
\begin{proof}
Let $\A$ be the original algorithm and denote the evaluation time by $T$. Our modified procedure instantiates $k = O(\log(1/\delta))$ independent copies $\A_1, \dots, \A_k$ of $\A$ to run in parallel, choosing one copy uniformly at random to use for each round. After sufficiently many rounds, it simply returns the median of the current estimates maintained by the copies. 

To begin our analysis, let $i_t \sim \Unif([k])$ denote the index of the copy used at time $t$, and, for each $i \in [k]$, take $S_i = \{ t \in [T] : i_t = i \}$ to be the set of times for which copy $i$ is chosen. Further, let $F^{(i)}_T$ denote the empirical cdf of the samples from the times in $S_i$. Now, we note that each $|S_i| \approx T/k$ for sufficiently large $T$. In particular, a Chernoff bound gives
\begin{equation}\label{eq: chernoff equal pieces}
    \Pr\left[\left||S_i| - \frac{T}{k}\right| > \eta \frac{T}{k}\right] \leq 2\exp\left(-\frac{\eta^2T}{3 k} \right) \leq \frac{\delta}{4 k}
\end{equation}
for $T \geq 3 k \log(8k/\delta)/\eta^2$. Next, we will show that the means and CDFs of the $S_i$ samples are both close to those for the full set of $T$ samples, since even an adaptive adversary has no hope of predicting the randomly chosen indices.

We fully describe the case of mean estimation, letting $\overline{x}_T^{(i)}$ denote the mean of the samples seen by copy $i$. In the case of an oblivious adversary, it is easy to check that $\Exp\bigl[ \sum_{t \in S_i} x_t \bigr] = \frac{T}{k} \overline{x}_T$. Thus, we can apply Chernoff once more to obtain
\begin{align}\label{eq: chernoff sum concentration}
    \Pr\left[\left|\sum_{t \in S_i} x_t - \frac{T}{k} \overline{x}_T \right| > \eta\frac{T}{k} \right] &= \Pr\left[\left|\sum_{t \in S_i} x_t - \frac{T}{k} \overline{x}_T \right| > \frac{\eta}{\overline{x}_T} \frac{T}{k} \overline{x}_T \right]\nonumber\\
    &\leq 2\exp\left(-\frac{\eta^2 T}{3 n k \,\overline{x}_T}\right) \leq 2\exp\left(-\frac{\eta^2 T}{3 n^2 k}\right) \leq \frac{\delta}{4k}
\end{align}
for $T \geq 3 k n^2 \log(8k/\delta) / \eta^2$. In the adaptive case, the values $x_t$ chosen by the adversary are dependent on previous queries, but we can reach an identical bound by considering martingale increments and applying the multiplicative version of Azuma's inequality for martingale increments.

Next, we exploit the relative smoothness of $(x,y) \mapsto x/y$, noting that, for $a,b \geq 0$,
\begin{equation*}
    \left|\frac{A}{B} - \frac{a}{b}\right| \leq \frac{b|A-a| + a|B - b|}{b^2 - b|B-b|}.
\end{equation*}
Setting $A = \sum_{t \in S_i} x_i$, $a = \frac{T}{k}\overline{x}_T \leq \frac{Tn}{k}$, $B = |S_i|$, $b = \frac{T}{k}$, and substituting $\eta = \frac{\eps}{6}$ for \eqref{eq: chernoff equal pieces} and $\eta = \frac{\eps n}{6}$ for \eqref{eq: chernoff sum concentration}, we obtain
\begin{equation*}
    \left|\overline{x}_T^{(i)} - \overline{x}_T\right| \leq \frac{\frac{T}{k}\left(\frac{n\eps}{6}\frac{T}{k}\right) + \frac{Tn}{k}\left(\frac{\eps}{6} \frac{T}{k} \right)}{\left(\frac{T}{k}\right)^2 - \frac{T}{k}\left(\frac{\eps}{6} \frac{T}{k}\right)} = \frac{\frac{\eps n}{6} + \frac{\eps n}{6}}{1 - \frac{\eps}{6}} \leq \frac{\eps n}{2}
\end{equation*}
for all $i \in [k]$, conditioned on an event with probability at least $1 - 2k\frac{\delta}{4k} = 1 - \delta/2$, so long as $T \geq 108 k \log(8k/\delta) / \eps^2$. Now, recall that the mean estimate $\hat{\mu}_T^{(i)}$ produced by copy $i$ is within $n\eps/2$ of $\overline{x}_T^{(i)}$ with probability at least $3/4$, so long as $|S_i| \geq T_0(n, \eps/2)$, which occurs under the same conditioning when, for example, $T \geq 2kT_0(n,\eps/2)$. In this case, fixing $k = \Theta(\log(1/\delta))$ ensures that the median of these estimates is $\eps$-good with respect to the entire sample set with probability at least $1 - \delta/2$. Unrolling the conditioning and examining all of our lower bounds on $T$, we find that an $\eps$-good mean is produced with probability at least $1-\delta$ so long as $T \geq \max\{T_0(n,\eps/2)\Theta(\log(1/\delta)), \Theta(\text{polylog}(1/\delta)/\eps^2)\}$.

The case of CDF estimation (which implies the result for median estimation) follows in much the same way,  where we must show that each $F_T^{(i)} \approx F_T$ with high probability for sufficiently large $T$. Specifically, one can show that $T \geq \max\{T_0(n,\eps/2)\Theta(\log(1/\delta)),\Theta(n\,\text{polylog}(1/\delta)/\eps^2)$ is sufficient, where the factor of $n$ appears because of a union bound over the $n$ values of the CDF function.
\end{proof}

\end{document}